\documentclass[twoside]{article}

\usepackage[accepted]{aistats2019}
\usepackage{times}

\usepackage[utf8]{inputenc}
\usepackage[T1]{fontenc} 
\usepackage[small]{caption}
\usepackage{soul}
\usepackage{hyperref}
\usepackage{nicefrac}
\usepackage{microtype}
\usepackage{wrapfig}
\usepackage{lipsum}

\usepackage{amsmath}
\usepackage{amssymb}
\usepackage{amstext}
\usepackage{amsfonts}

\usepackage{bm}
\usepackage{acronym}
\usepackage{float}
\usepackage{tikz}
\usetikzlibrary{bayesnet}
\usetikzlibrary{shapes.multipart}
\usepackage{pgfplots}
\usepackage[linesnumbered,ruled]{algorithm2e}
\usepackage{graphicx}
\usepackage{subcaption}
\usepackage{booktabs}
\usepackage[inline]{enumitem}
\usepackage{todonotes}
\usepackage{hyperref}
\usepackage{cleveref}
\crefname{algocf}{alg.}{algs.}
\Crefname{algocf}{Algorithm}{Algorithms}
\crefname{enumi}{}{}

\crefname{equation}{eq}{eqs}
\Crefname{equation}{Eq}{Eqs}
\crefname{table}{table}{tables}
\Crefname{table}{Table}{Tables}

\crefname{figure}{fig}{figs}
\Crefname{figure}{Fig}{Figs}

\acrodef{IRT}{Item Response Theory}
\acrodef{ICC}{Item Characteristic Curve}
\acrodef{ELBO}{Evidence Lower Bound}
\acrodef{ML}{Machine Learning}
\acrodef{KL}{Kullback–Leibler}
\acrodef{SD}{Standard Deviation}
\acrodef{MLP}{Multi-layer Perceptron}
\acrodef{KNN}{k-Nearest Neighbours}
\acrodef{LDA}{Linear Discriminant Analysis}
\acrodef{QDA}{Quadratic Discriminant Analysis}
\acrodef{CDF}{Cumulative Distribution Function}
\acrodef{AUC}{Area Under the ROC-Curve}
\acrodef{VI}{Variational Inference}
\acrodef{LR}{Logistic Regression}
\acrodef{SGD}{Stochastic Gradient Descent}
\acrodef{MLE}{Maximum Likelihood}

\newcommand{\BIRT}{$\beta^3$-\ac{IRT}} 

\newcommand{\high}{{\color{red}high}}
\newcommand{\low}{{\color{green}low}}

\newcommand{\Beta}{{\mathcal{B}}}

\newcommand{\Bernoulli}{{\mathcal{B}ern}}

\renewcommand{\a}{{\mathbf{a}}}
\newcommand{\deltab}{{\bm{\delta}}}
\newcommand{\thetab}{{\bm{\theta}}}

\newcommand{\Fcal}{{\mathcal{F}}}

\newcommand{\EE}{{\mathbb{E}}}
\newcommand{\KL}{{\mathbb{KL}}}
\newcommand{\Ind}{{\mathbb{I}}}

\newcommand{\MNIST}{{\textsc{mnist}}}
\newcommand{\clusters}{{\textsc{clusters}}}
\newcommand{\moons}{{\textsc{moons}}}

\begin{document}

%

%
\renewcommand*{\thefootnote}{\fnsymbol{footnote}}

\twocolumn[

\aistatstitle{$\bm{\beta}^3$-IRT: A New Item Response Model and its Applications}

\aistatsauthor{ Yu Chen \And Telmo Silva Filho} 
\aistatsaddress{University of Bristol \And  Universidade Federal de Pernambuco}
\aistatsauthor{Ricardo B. C. Prud\^encio \And Tom Diethe \And Peter Flach} 
\aistatsaddress{Universidade Federal de Pernambuco \And Amazon \And University of Bristol \& Alan Turing Institute}
]

\begin{abstract}
\acf{IRT} aims to assess latent abilities of respondents based on the correctness of their answers in aptitude test items with different difficulty levels.  
In this paper, we propose the 
\BIRT{} model, 
which models continuous responses and 
can generate a much enriched family of \aclp*{ICC}. 
In experiments we applied
the proposed model to data from an online exam platform, and show our model outperforms a more standard 2PL-ND model on all datasets. Furthermore, we show how to apply \BIRT{} to assess the ability of machine learning classifiers. This novel application results in a new metric for evaluating the quality of the classifier's probability estimates, based on the inferred difficulty and discrimination of data instances.  



\end{abstract}


\renewcommand*{\thefootnote}{\arabic{footnote}}

\section{INTRODUCTION}


\label{sec:introduction}

\ac{IRT} is widely adopted in psychometrics for estimating human latent ability in tests. Unlike classical test theory, which is concerned with performance at the test level, IRT focuses on the items, modelling responses given by respondents with different abilities to items of different difficulties, both measured in a known scale \cite{embretson2013item}. The concept of item depends on the application, and can represent for instance test questions, judgements or choices in exams. In practice, IRT models estimate latent difficulties of the items and the latent abilities of the respondents based on observed responses in a test, and have been commonly applied to assess performance of students in exams.

Recently, IRT was adopted to analyse \ac{ML} classification tasks \cite{martinez2016making}. Here, items correspond to instances in a dataset, while respondents are classifiers. The responses are the outcomes of classifiers on the test instances (right or wrong decisions collected in a cross-validation experiment). Our work is different in two aspects: 1) we propose a new IRT model for continuous responses; 2) we applied the proposed IRT to predicted probabilities instead of binary responses.

For each item (instance), an \acf{ICC} is estimated, which is a logistic function that returns the probability of a correct response for the item based on the respondent ability. This \ac{ICC} is determined by two item parameters: difficulty, which is the location parameter of the logistic function; and discrimination, which affects the slope of the ICC. 
Despite the useful insights \cite{martinez2016making} and applicability to other AI contexts, binary IRT models are limited when the techniques of interest return continuous responses (e.g. class probabilities). Continuous IRT models have been developed and applied in psychometrics \cite{noel2007beta} but have limitations when applied in this context: 
limited interpretability since abilities and difficulties are expressed as real numbers; and 
limited flexibility since \acp{ICC} are limited to logistic functions. 

In this paper, we propose a novel \ac{IRT} model called \BIRT{} to addresses these limitations by means of a new parameterisation of \ac{IRT} models such that: 
\begin{enumerate*}[label=\alph*\upshape)]
    \item the resulting \acp{ICC} are not limited to logistic curves, different shapes can be obtained depending on the item parameters, which allows  more flexibility when fitting responses for different items;
    \item abilities and difficulties are in the $[0,1]$ range, which gives a unified scale for easier interpretation and evaluation.
\end{enumerate*}


%
We first applied the \BIRT{} model  in a case study to predict students' responses in an online education platform. Out model outperforms the 2PL-ND model \cite{noel2007beta} on all datasets. 
In a second case study, \BIRT{} was used to fit class probabilities estimated by classifiers in binary classification tasks. The experimental results show the ability inferred by the model allows to evaluate probability estimation of classifiers in an instance-wise manner with the aid of item parameters (difficulty and discrimination). 
Hence we show that the \BIRT{} model is useful not only in the more traditional applications associated with \ac{IRT}, but also in the \ac{ML} context outlined above.


  {Our contributions can be summarised as follows:
    \begin{itemize}
    \item we propose a new model for \ac{IRT} with richer \ac{ICC} shapes, which is more versatile than existing models;
    \item we demonstrate empirically that this model has better predictive power; 
    \item we demonstrate the use of this model in an \ac{ML} setting, providing a method for assessing the `ability' of classification algorithms that is based on an instance-wise metric in terms of class probability estimation. 
    \end{itemize}
  }

The paper is organised as follows. \Cref{sec:binary:irt} gives a brief introduction of Binary \acf{IRT}. \Cref{sec:birt} presents the \BIRT{} model, followed by related work on IRT in \Cref{sec:related}. \Cref{sec:experiments:students} presents the experiments on real students, while \Cref{sec:experiments:ml} presents the use of \BIRT{} to evaluate classifiers. Finally, \Cref{sec:conclusion} concludes the paper. 



\section{BINARY ITEM-RESPONSE THEORY}
\label{sec:binary:irt}


In \acl{IRT}, the probability of a correct response for an item depends on the latent respondent ability and the item difficulty. Most previous studies on \ac{IRT} have adopted binary models, in which the responses are either correct or incorrect. 
%
%
Such models assume a binary response $x_{ij}$ of the $i$-th respondent to the $j$-th item. In the IRT model with two item parameters (2PL), the probability of a correct response ($x_{ij} = 1$) is defined by a logistic function with location parameter $\deltab_j$ and shape parameter $\a_j$. Responses are modelled by the Bernoulli distribution with parameter $p_{ij}$ as follows: 
\begin{equation} 
\label{eq:genIRT}
    x_{ij} = \Bernoulli(p_{ij}),\; 
    p_{ij} = \sigma(-{\a_j} d_{ij}),\;
    d_{ij} = {\thetab_i-\deltab_j}\\ 
\end{equation}
 where $\sigma(\cdot)$ is the logistic function. 
$N$ is the number of items and $M$ is the number of participants.

The 2PL model gives a logistic \acf{ICC} mapping ability $\thetab_i$ to expected response 
as follows: 
\begin{equation}
\label{eq:2plIRT}
\EE[x_{ij}|\thetab_i,\deltab_j,\a_j] = p_{ij} = \frac{1}{1+e^{-\a_j(\thetab_i-\deltab_j)}}
\end{equation}
%
At $\thetab_i = \deltab_j$ the expected response is $0.5$. 
The slope of the ICC at $\thetab_i = \deltab_j$
 is $\a_j/4$. If $\a_j=1, \forall j=1,\ldots,N$, a simpler model is obtained, known as 1PL, which describes items solely by their difficulties. Generally,
discrimination $\a_j$ indicates how probability of correct responses changes as the ability increases. High discriminations lead to steep ICCs at the point where ability equals difficulty, with small changes in ability producing significant changes in the probability of correct response.

The standard IRT model \emph{implicitly} uses a noise model that follows a logistic distribution (in the same way that logistic regression does). This can be replaced with Normally distributed noise, which results in the logistic link function being replaced by a probit link function (Gaussian \acf{CDF}), as is the case in the DARE model \cite{bachrach2012grade}. In practice, the probit link function is very similar to the logistic one, so these two models tend to behave very similarly, and the particular choice is often then based on mathematical convenience or computation tractability.

Despite their extensive use in psychometrics, binary \ac{IRT} models are of limited use when responses are naturally produced in continuous scales. Particularly in \ac{ML}, binary models are not adequate if the responses to evaluate are class probability estimates. 

\section{THE \texorpdfstring{\BIRT{}}~ MODEL}\label{sec:birt}

 \begin{figure}[!t]
 \centering
     \includegraphics[width=0.7\linewidth,trim={5.2cm 16.cm 9.5cm 4.5cm},clip]{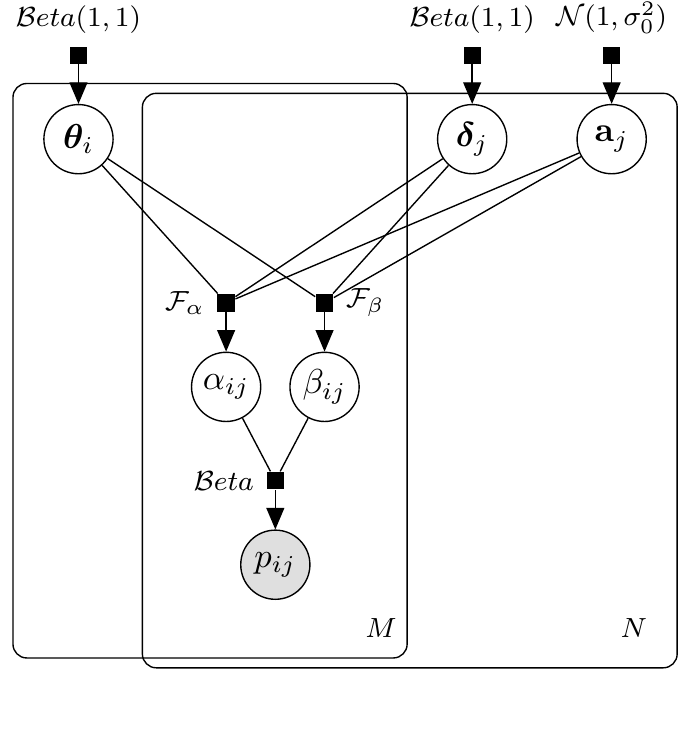}
\caption{Factor Graph of \BIRT{} Model. The grey circle represents observed data, white circles represent latent variables, small black rectangles represent stochastic factors, and the rectangular plates represent replicates.}
\label{fig:birt}
 \end{figure}

\begin{figure*}[!tb]
\centering
\begin{subfigure}{0.3\linewidth}
\includegraphics[width=\linewidth,trim={1cm 0.cm 0.cm 0.cm}]{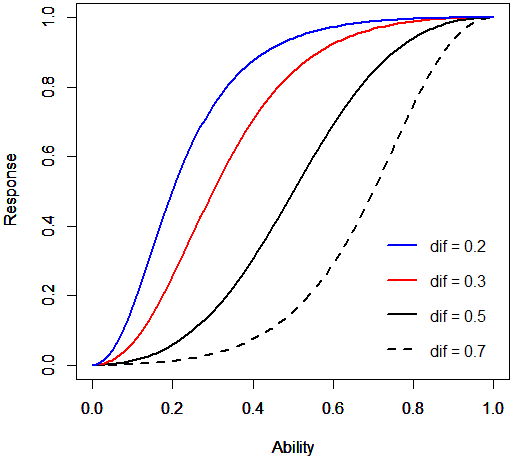}
\caption{$\a_j = 2$.}
\label{fig:ICC_discr2}
\end{subfigure}
\quad
\begin{subfigure}{0.3\linewidth}
\includegraphics[width=\linewidth,trim={1cm 0.cm 0.cm 0.cm}]{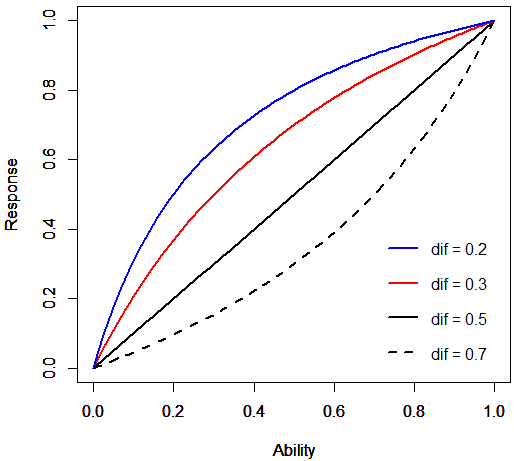}
\caption{$\a_j = 1$.}
\label{fig:ICC_discr1}
\end{subfigure}
\quad
\begin{subfigure}{0.3\linewidth}
\includegraphics[width=\linewidth,trim={1cm 0.cm 0.cm 0.cm}]{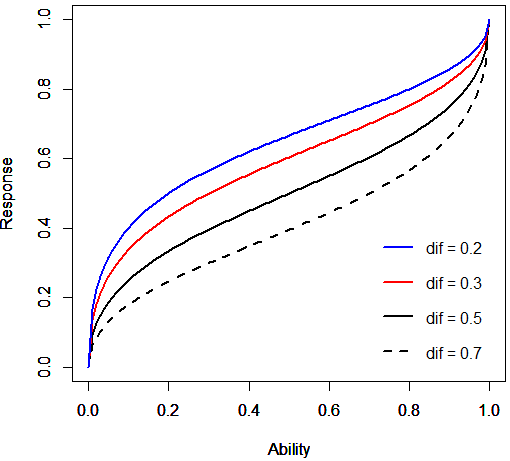}
\caption{$\a_j = 0.5$.}
\label{fig:ICC_discr05}
\end{subfigure}
%
\caption{Examples of Beta ICCs for Different Values of Difficulty and Discrimination. Higher discriminations lead to steeper ICCs, higher difficulties need higher abilities to achieve higher responses.}
\label{fig:betaICCs}
\end{figure*}
%
%

In this Section, we will elaborate on the parametrisation and inference method of \BIRT{} model, highlighting the differences with existing \ac{IRT} models.

\subsection{Model description}



The factor graph of \BIRT{} is shown in \Cref{fig:birt} and \Cref{eq:model_def} below gives the model definition, where $M$ is number of respondents and $N$ is number of items, $p_{ij}$ is the observed response of respondent $i$ to item $j$, which is drawn from a Beta distribution:
\begin{align} \label{eq:model_def}
    p_{ij} &\sim \Beta(\alpha_{ij}, \beta_{ij}), \notag \\
    \alpha_{ij} &= \Fcal_{\alpha}(\thetab_i, \deltab_j, \a_j) = 
    \left(\frac{\thetab_i}{\deltab_j}\right)^{\a_j}, \notag \\
    \beta_{ij} &= \Fcal_{\beta}(\thetab_i, \deltab_j, \a_j) = 
    \left(\frac{1-\thetab_i}{1-\deltab_j}\right)^{\a_j}, \notag \\
    \thetab_i &\sim \Beta(1,1),\;
    \deltab_j \sim \Beta(1,1),\;
    \a_j \sim \mathcal{N}(1, \sigma^2_{0})
\end{align}
%
%
%
The parameters $\alpha_{ij},\beta_{ij}$ are computed from $\thetab_i$ (the ability of participant $i$), $\deltab_j$ (the difficulty of item $j$), and $\a_j$ (the discrimination of item $j$). $\thetab_i$ and $\deltab_j$ are also drawn from Beta distributions and their priors are set to $\Beta(1,1)$ in a general setting. 
The discrimination $\a_j$ is drawn from a Normal distribution with prior mean $1$ and variance $\sigma^2_0$, where 
$\sigma_0$ is a hyperparameter of the model. These uninformative priors are used as default setting of the model, but could be parametrised by further hyperparameters when there is further prior information available.
The default prior mean of $\a_j$ is set to $1$ rather than $0$ because the discrimination $\a_j$ is a power factor here.

 When comparing probabilities we use ratios (e.g. the likelihood ratio). Similarly, here we use the ratio of ability to difficulty since in our model these are measured on a $[0,1]$ scale as well: a ratio smaller/larger than 1 means that ability is lower/higher than difficulty. These ratios are positive reals and hence map directly to $\alpha$ and $\beta$ in \Cref{eq:model_def}. Importantly, the new parametrisation enables us to obtain non-logistic \acp{ICC}.
In this model the \ac{ICC} is defined by the expectation of $\Beta(\alpha_{ij}, \beta_{ij})$ and then assumes the form:
\begin{equation}\label{eq:ep_birt}
    \begin{split}
        \EE[p_{ij}|\thetab_i,\deltab_j,\a_j] &= \frac{\alpha_{ij}}{\alpha_{ij}+\beta_{ij}} 
        = \frac{1}{1+\left(\frac{\deltab_j}{1-\deltab_j}\right)^{\a_j} \left(\frac{\thetab_i}{1-\thetab_i}\right)^{-\a_j}}
    \end{split}
\end{equation}
%
%
As in standard IRT, the difficulty $\deltab_j$ is a location parameter. The response is 0.5 when $\thetab_i=\deltab_j$ and the curve has slope $\a_j/(4\deltab_j(1-\deltab_j))$ at that point. \Cref{fig:betaICCs} shows examples of Beta \acp{ICC} for different regimes depending on $\a_j$:
%
\begin{align*}
        &\a_j > 1: &\; &\text{a sigmoid shape similar to stardard IRT}, \\
        &\a_j = 1: &&\text{parabolic curves with vertex at 0.5}, \\
    0 < &\a_j < 1: &&\text{anti-sigmoidal behaviour}.
%
\intertext{%
Note that the model allows for negative discrimination, which indicates items that are somehow harder for more able respondents. 
Negative $\a_j$ can be divided similarly: 
}%
    -1 < &\a_j < 0: &\; &\text{decreasing anti-sigmoid}, \\
         &\a_j < 1: &&\text{decreasing sigmoid}.
\end{align*}
We will see examples in \Cref{sec:experiments:ml} that negative discrimination can in fact be useful for identifying `noisy' items, where higher abilities getting lower responses. 
%
%
%
\subsection{Model inference}
%


We tested two inference methods on \BIRT{} model, one is conventional \ac{MLE} which we applied to experiments with student answers (\Cref{sec:experiments:students}) for response prediction, using the likelihood function shown in \Cref{eq:ep_birt}.

The other is Bayesian \ac{VI} \cite{bishop2006pattern} which we applied to experiments with classifiers (\Cref{sec:experiments:ml}) for full Bayesian inference on latent variables. 
In \ac{VI}, the optimisation objective is the lower bound of \ac{KL} divergence between the true posterior and variational posterior of latent variables, which is also referred as \ac{ELBO}. However, the model is highly non-identifiable because of its symmetry \cite{nishihara2013detecting}, resulting in undesirable combinations of these variables: for instance, when $p_{ij}$ is close to $1$, it usually indicates $\alpha_{ij} > 1$ and $\beta_{ij} < 1$, which can arise when $\thetab_i > \deltab_j$ with positive $\a_j$, or $\thetab_i < \deltab_j$ with negative $\a_j$. Hence, we update discrimination as a global variable after ability and difficulty converge at each step (see also \Cref{alg:vi}), in addition to setting the prior of discrimination as N(1,1) to reflect the assumption that discrimination is more often positive than negative. We employ the coordinate ascent method of \ac{VI} \cite{blei2017variational}, keeping $\a_j$ fixed while optimising $\thetab_i$ and $\deltab_j$ and vice versa. 
Accordingly, the two separate loss functions are defined as below:
\begin{align}\label{eq:elbo_local}
     \mathcal{L}_1 &= \sum_{i=1}^M \sum_{j=1}^N \EE_{q}[\log p(p_{ij}|\thetab_i,\deltab_j)] \notag \\
     &+ \sum_{i=1}^M \EE_q[ \log p(\thetab_i) - \log q(\thetab_i|\bm{\phi}_i)] \notag \\
     &+ \sum_{j=1}^N \EE_q[\log p(\deltab_j) - \log q(\deltab_j|\bm{\psi}_j)]
\end{align}
\begin{equation}\label{eq:elbo_global}
        \mathcal{L}_{2} =  \sum_{j=1}^N \EE_{q}[\log p(p_{ij}|\a_{j}) + \log p(\a_j) - \log q(\a_j|\bm{\lambda}_j)]  
\end{equation}
Here, $\bm{\phi},\bm{\psi},\bm{\lambda}$ are parameters of variational posteriors of $\thetab, \deltab, \a$, respectively. $\mathcal{L}_1$ and $\mathcal{L}_2$ are lower bounds of $\KL(p(\thetab,\deltab|\mathbf{x})||q(\thetab,\deltab|\bm{\phi},\bm{\psi}))$ and $\KL(p(\a|\mathbf{x})||q(\a|\bm{\lambda})$, respectively. Both are optimised using \ac{SGD}. 
\begin{algorithm}[!t]
    Set number of iterations $L_{iter}$,
    randomly initialise $\bm{\phi}, \bm{\psi}$, $\bm{\lambda}$\;
    \For{$t$ in range($L_{iter}$)}
    {
        \While{$\Theta = \{\bm{\phi},\bm{\psi}\}$ not converged}
        {
            Compute $\nabla_{\Theta} \mathcal{L}_1$ according to \Cref{eq:elbo_local};\\
            Update $\Theta$ by \ac{SGD};
        }
        
        Compute $\nabla_{\bm{\lambda}} \mathcal{L}_2$ according to \Cref{eq:elbo_global};\\
            Update $\bm{\lambda}$ by \ac{SGD};
        
    }
 \caption{Variational Inference for \BIRT{}}
 \label{alg:vi}
\end{algorithm}
In order to apply the reparameterisation trick \cite{kingma2013auto}, we use Logit-Normal to approximate the Beta distribution in variational posteriors. The steps to perform \ac{VI} of the model are shown in \Cref{alg:vi}, where the variational parameters can be updated by any gradient descent optimisation algorithm (we use the Adam method \cite{kingma2014adam} in our experiments). We implemented \Cref{alg:vi} using the probabilistic programming library Edward \cite{tran2016edward}.

\subsection{Related work}
\label{sec:related}

There have been earlier approaches to IRT with continuous approaches. In particular, 
\cite{noel2007beta} proposed an IRT model which adopts the Beta distribution with parameters $m_{ij}$ and $n_{ij}$ as follows:  
%
\begin{align*} 
    m_{ij} &= e^{\left(\thetab_i-\deltab_j\right)/2}, \quad
    n_{ij} = 
    e^{-\left(\thetab_i-\deltab_j\right)/2} = m_{ij}^{-1}, \\
    p_{ij} &\sim \Beta(m_{ij}, n_{ij}).
\end{align*}
%
This model gives a logistic \ac{ICC} mapping ability to expected response for item $j$ of the form: 
\begin{equation}
    \label{eq:1plNoel}
    \EE[p_{ij}|\thetab_i,\deltab_j] =  \frac{m_{ij}}{m_{ij}+n_{ij}} = \frac{1}{1+e^{-(\thetab_i-\deltab_j)}}
\end{equation}
%
%
While there is a superficial similarity to the \BIRT{} model, there are two crucial distinctions. 
\begin{itemize}
    \item Similarly to the standard 1PL IRT model, \Cref{eq:1plNoel} does not have a discrimination parameter. The ICC therefore has a fixed slope of $0.25$ at $\thetab_i=\deltab_j$ and it is assumed that all items have the same discrimination. 
    \item \Cref{eq:1plNoel} assumes a real-valued scale for abilities and difficulties, whereas \BIRT{} uses a $[0,1]$ scale. Not only does this avoid problems with interpreting extreme values, but more importantly it opens the door to the non-sigmoidal \acp{ICC} as depicted in Figure~\ref{fig:betaICCs}.
\end{itemize}

\section{EXPERIMENTS WITH STUDENT ANSWER DATASETS}
\label{sec:experiments:students}

We begin by applying and evaluating \BIRT{} to model responses of students, which is a common application of IRT. We use datasets that consist of answers given by students of $33$ different courses from an online platform (which we cannot disclose, for commercial reasons). The courses have different numbers of questions, but not all students have answered all questions for a given course (in fact, most did not), therefore we did not have $p_{ij}$ values for every student $i$ and question $j$. On the other hand, it is possible for a student to answer a question multiple times. These datasets may contain noise derived from user behaviour, such as quickly answering questions just to see the next ones and lending accounts to other students.

We compare the \BIRT{} model with Noel and Dauvier's continuous IRT model. 
Since without a discrimination parameter their model is rather weak, we strengthen it by introducing a discrimination parameter as follows: 
\begin{equation}
    \label{eq:2plNoel}
    \EE[p_{ij}|\thetab_i,\deltab_j,\a_j] = \frac{m_{ij}}{m_{ij}+n_{ij}} = \frac{1}{1+e^{-\a_j(\thetab_i-\deltab_j)}}
\end{equation}
We refer to this model as 2PL-ND. 

Our experiments follow two scenarios. In the first scenario, to generate the continuous responses, we take the $i$-th student's average performance for the $j$-th question, $p_{ij} = \EE[x_{ij}]$. In the second scenario, only the first attempts of each student for each question are considered, leading to a binary problem.  
The models were trained by minimising the log-loss of predicted probabilities using \ac{SGD}, implemented in Python, with the Theano library. We trained the models for $2500$ iterations, using batches of $2000$ answers and an adaptive learning rate calculated as $0.5 / \sqrt{t}$, where $t$ is the current iteration. \BIRT{} predicts the probability that the $i$-th student will answer the $j$-th question correctly following \Cref{eq:ep_birt}, while 2PL-ND follows \Cref{eq:2plNoel}.

For 2PL-ND, abilities and difficulties are unbounded and drawn from $\mathcal{N}(0, 1)$. For \BIRT{}, abilities are bounded in the range $(0, 1)$, so they were drawn from $\mathcal{B}(m^+_i, m^-_i)$, were  $m^+_i$ ($m^-_i$) represent the number of correct (incorrect) answers given by the $i$-th student. Likewise, difficulties were drawn from $\mathcal{B}(n^+_j,n^-_j)$, where  $n^+_j$ ($n^-_j$) represent the number of correct (incorrect) answers given to the $j$-th question. For both models, discriminations were drawn from $\mathcal{N}(1, 1)$.

In the experiments, we ran $30$ hold-out schemes with $90\%$ of the dataset kept for training and $10\%$ for testing. For a training set to be valid, every student and question must occur in it at least once. Therefore, we sample the training and test sets with stratification based on the students and then we check which questions are absent, sampling $90\%$ of their answers for training and $10\%$ for testing. 

\begin{figure}
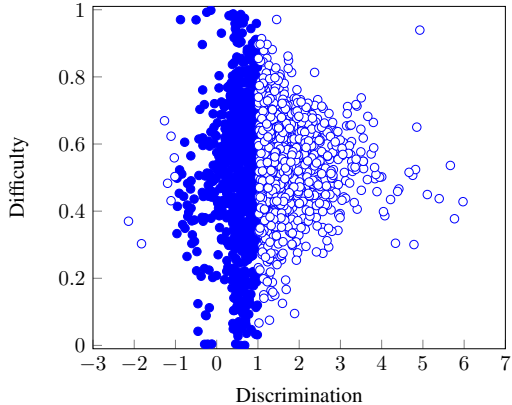

    \centering


    \captionof{figure}{Discrimination versus Difficulty (Inferred in one of the $30$ hold-out schemes across all $33$ student answer datasets. Solid dots represent discriminations in the $(-1, 1)$ range, which produce non-logistic curves in our approach. These represent approximately $46\%$ of all estimated discriminations.)}
    \label{fig:disc:vs:diff}
\end{figure}

\begin{table}[!tb]
     \tiny
     \centering
         \caption{Student Answer Datasets (Log-loss) for Continuous (Student's Average) and First Attempt Performance.
        }
     \label{tab:disciplines}
     \begin{tabular}{ccccc}
\toprule
         &                    \multicolumn{2}{c}{continuous}                  & \multicolumn{2}{c}{first attempts}\\
\midrule
course   &      
$\bm{\beta}^3$-IRT                    &  2PL-ND            &   $\bm{\beta}^3$-IRT                    &  2PL-ND            \\
\midrule
 1            &      \textbf{0.631 $\pm$ 0.003} &  0.713 $\pm$ 0.004 & \textbf{0.623 $\pm$ 0.004} &  0.699 $\pm$ 0.005 \\
 2             &   \textbf{0.630 $\pm$ 0.022} &  0.972 $\pm$ 0.081 & \textbf{0.623 $\pm$ 0.023} &  0.953 $\pm$ 0.060 \\
 3            &    \textbf{0.617 $\pm$ 0.004} &  0.695 $\pm$ 0.004 & \textbf{0.628 $\pm$ 0.024} &  0.760 $\pm$ 0.086 \\
 4             &    \textbf{0.671 $\pm$ 0.004} &  0.742 $\pm$ 0.009 & \textbf{0.669 $\pm$ 0.004} &  0.731 $\pm$ 0.007 \\
 5             &    \textbf{0.594 $\pm$ 0.004} &  0.692 $\pm$ 0.008 & \textbf{0.597 $\pm$ 0.004} &  0.696 $\pm$ 0.013 \\
 6             &    \textbf{0.661 $\pm$ 0.009} &  0.899 $\pm$ 0.039 & \textbf{0.651 $\pm$ 0.009} &  0.892 $\pm$ 0.030 \\
 7             &    \textbf{0.630 $\pm$ 0.007} &  0.795 $\pm$ 0.020 & \textbf{0.632 $\pm$ 0.007} &  0.791 $\pm$ 0.015 \\
 8              &     \textbf{0.648 $\pm$ 0.014} &  0.941 $\pm$ 0.044 & \textbf{0.641 $\pm$ 0.023} &  0.967 $\pm$ 0.059 \\
 9             &     \textbf{0.657 $\pm$ 0.011} &  0.941 $\pm$ 0.030 & \textbf{0.660 $\pm$ 0.011} &  0.931 $\pm$ 0.032 \\
10             &    \textbf{0.649 $\pm$ 0.007} &  0.847 $\pm$ 0.030 & \textbf{0.655 $\pm$ 0.009} &  0.841 $\pm$ 0.032 \\
11             &    \textbf{0.633 $\pm$ 0.016} &  0.889 $\pm$ 0.051 & \textbf{0.630 $\pm$ 0.012} &  0.891 $\pm$ 0.067 \\
12             &   \textbf{0.650 $\pm$ 0.013} &  0.938 $\pm$ 0.063 & \textbf{0.662 $\pm$ 0.016} &  0.883 $\pm$ 0.051 \\
13           &      \textbf{0.697 $\pm$ 0.066} &  1.002 $\pm$ 0.218 & \textbf{0.659 $\pm$ 0.086} &  1.023 $\pm$ 0.423 \\
14              &   \textbf{0.642 $\pm$ 0.028} &  0.936 $\pm$ 0.074 & \textbf{0.623 $\pm$ 0.028} &  0.909 $\pm$ 0.057 \\
15            &   \textbf{0.588 $\pm$ 0.002} &  0.650 $\pm$ 0.003 & \textbf{0.584 $\pm$ 0.003} &  0.642 $\pm$ 0.003 \\
16             &     \textbf{0.605 $\pm$ 0.002} &  0.674 $\pm$ 0.003 & \textbf{0.603 $\pm$ 0.002} &  0.663 $\pm$ 0.002 \\
17             &   \textbf{0.603 $\pm$ 0.002} &  0.665 $\pm$ 0.003 & \textbf{0.596 $\pm$ 0.003} &  0.657 $\pm$ 0.003 \\
18           &    \textbf{0.598 $\pm$ 0.006} &  0.725 $\pm$ 0.008 & \textbf{0.608 $\pm$ 0.005} &  0.729 $\pm$ 0.011 \\
19             &    \textbf{0.651 $\pm$ 0.015} &  0.923 $\pm$ 0.064 & \textbf{0.644 $\pm$ 0.020} &  0.934 $\pm$ 0.074 \\
20            &     \textbf{0.640 $\pm$ 0.021} &  0.959 $\pm$ 0.060 & \textbf{0.636 $\pm$ 0.018} &  0.933 $\pm$ 0.040 \\
21           &    \textbf{0.639 $\pm$ 0.016} &  0.949 $\pm$ 0.072 & \textbf{0.650 $\pm$ 0.014} &  0.968 $\pm$ 0.094 \\
22             &     \textbf{0.629 $\pm$ 0.020} &  0.935 $\pm$ 0.050 & \textbf{0.622 $\pm$ 0.016} &  0.931 $\pm$ 0.060 \\
23            &   \textbf{0.602 $\pm$ 0.004} &  0.692 $\pm$ 0.011 & \textbf{0.609 $\pm$ 0.004} &  0.682 $\pm$ 0.005 \\
24            &     \textbf{0.657 $\pm$ 0.014} &  0.950 $\pm$ 0.046 & \textbf{0.652 $\pm$ 0.011} &  0.950 $\pm$ 0.044 \\
25           &    \textbf{0.642 $\pm$ 0.015} &  0.917 $\pm$ 0.034 & \textbf{0.627 $\pm$ 0.010} &  0.871 $\pm$ 0.038 \\
26           &     \textbf{0.572 $\pm$ 0.011} &  0.836 $\pm$ 0.045 & \textbf{0.593 $\pm$ 0.014} &  0.874 $\pm$ 0.038 \\
27          &     \textbf{0.662 $\pm$ 0.022} &  0.998 $\pm$ 0.093 & \textbf{0.647 $\pm$ 0.028} &  0.971 $\pm$ 0.073 \\
28            &   \textbf{0.603 $\pm$ 0.001} &  0.647 $\pm$ 0.002 & \textbf{0.603 $\pm$ 0.001} &  0.645 $\pm$ 0.002 \\
29          &   \textbf{0.553 $\pm$ 0.021} &  0.916 $\pm$ 0.056 & \textbf{0.558 $\pm$ 0.017} &  0.856 $\pm$ 0.051 \\
30           &     \textbf{0.646 $\pm$ 0.019} &  1.001 $\pm$ 0.075 & \textbf{0.647 $\pm$ 0.019} &  0.979 $\pm$ 0.048 \\
31          &     \textbf{0.647 $\pm$ 0.014} &  0.911 $\pm$ 0.060 & \textbf{0.634 $\pm$ 0.014} &  0.918 $\pm$ 0.053 \\
32           &   \textbf{0.578 $\pm$ 0.001} &  0.627 $\pm$ 0.001 & \textbf{0.578 $\pm$ 0.002} &  0.626 $\pm$ 0.002 \\
33             &     \textbf{0.663 $\pm$ 0.021} &  0.929 $\pm$ 0.060 & \textbf{0.674 $\pm$ 0.025} &  0.993 $\pm$ 0.074 \\

\bottomrule
\end{tabular}
     \vspace{-0.3cm}
 \end{table}



\Cref{tab:disciplines} shows the log-loss for all 33 courses (full details of these courses can be found in the supplementary material). Best results were marked in bold, in case they were statistically significant according to a paired Wilcoxon signed-rank test, with significance level $0.001$. The results show that \BIRT{} outperformed 2PL-ND in all cases. We posit that this can be explained by the versatility of the model, which is able to find non-logistic \acp{ICC} when discriminations are taken from $(-1, 1)$.  \Cref{fig:disc:vs:diff} shows that roughly $46\%$ of all questions from the $33$ datasets are estimated to have discriminations in this interval, supporting this conclusion. As shown in \Cref{fig:ICC_discr05}, non-logistic \acp{ICC} are better at discriminating lower and higher ability values than logistic \acp{ICC}, which focus more on ability values around the corresponding question difficulty, and \BIRT{} is able to model both cases. Additionally, approximately $10\%$ of all questions had negative discriminations. One possible explanation is the presence of noise in these datasets, as mentioned earlier in this section. 
We will return to the relationship between negative discrimination and noise in \Cref{sec:experiments:ml}.



\section{\texorpdfstring{\BIRT{}}~ FOR CLASSIFIERS}
\label{sec:experiments:ml}

\begin{figure*}[!t]
    \centering
    \begin{subfigure}{0.48\linewidth}
        \includegraphics[width=\linewidth,trim={3.3cm 1.2cm 3.1cm 1.1cm},clip]{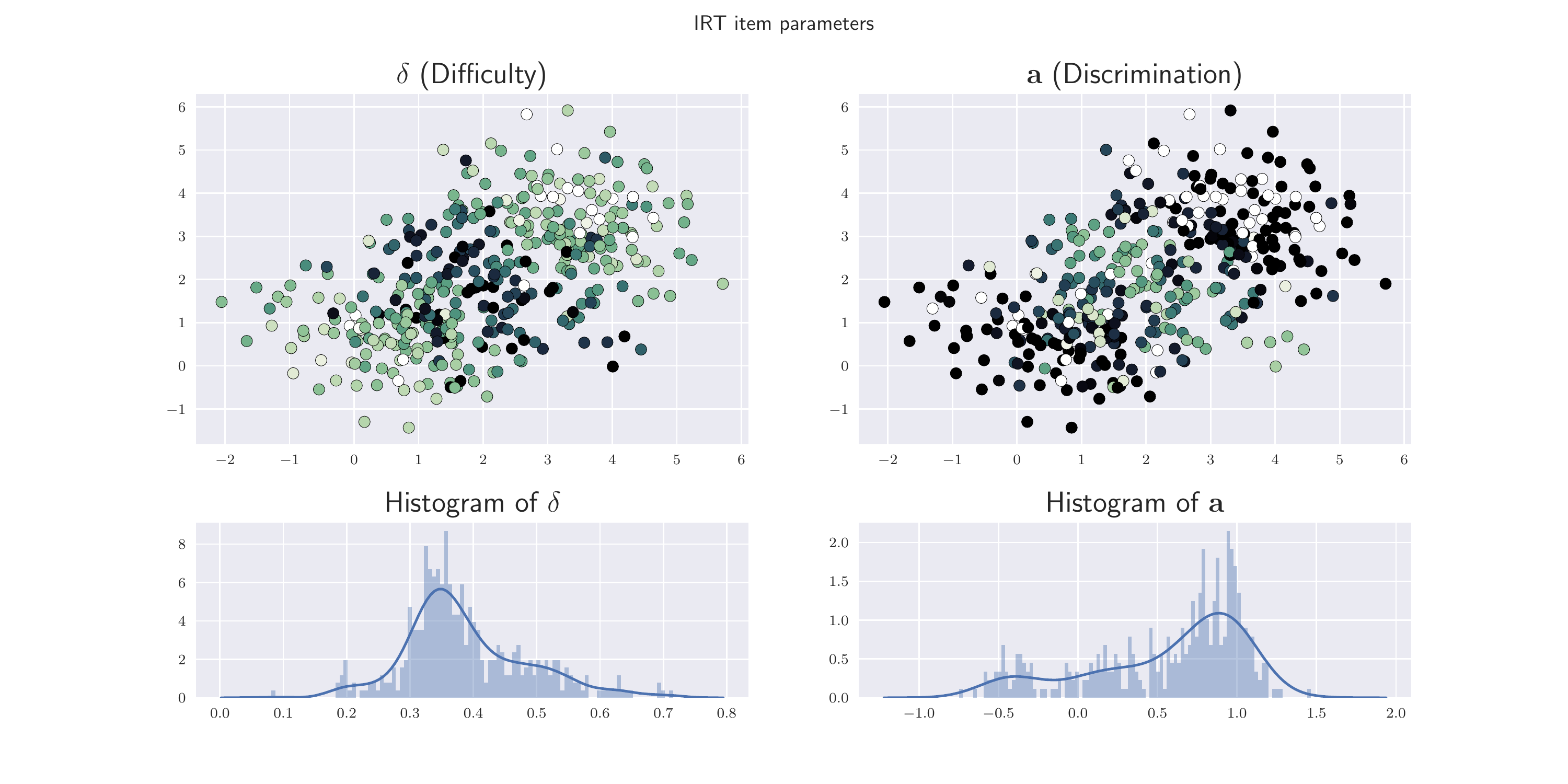}
    \caption{Dataset \clusters{}}
    \label{fig:cluster_r1}
    \end{subfigure}
    \quad
    \begin{subfigure}{0.48\linewidth}
        \includegraphics[width=\linewidth,trim={3.3cm 1.2cm 3.1cm 1.1cm},clip]{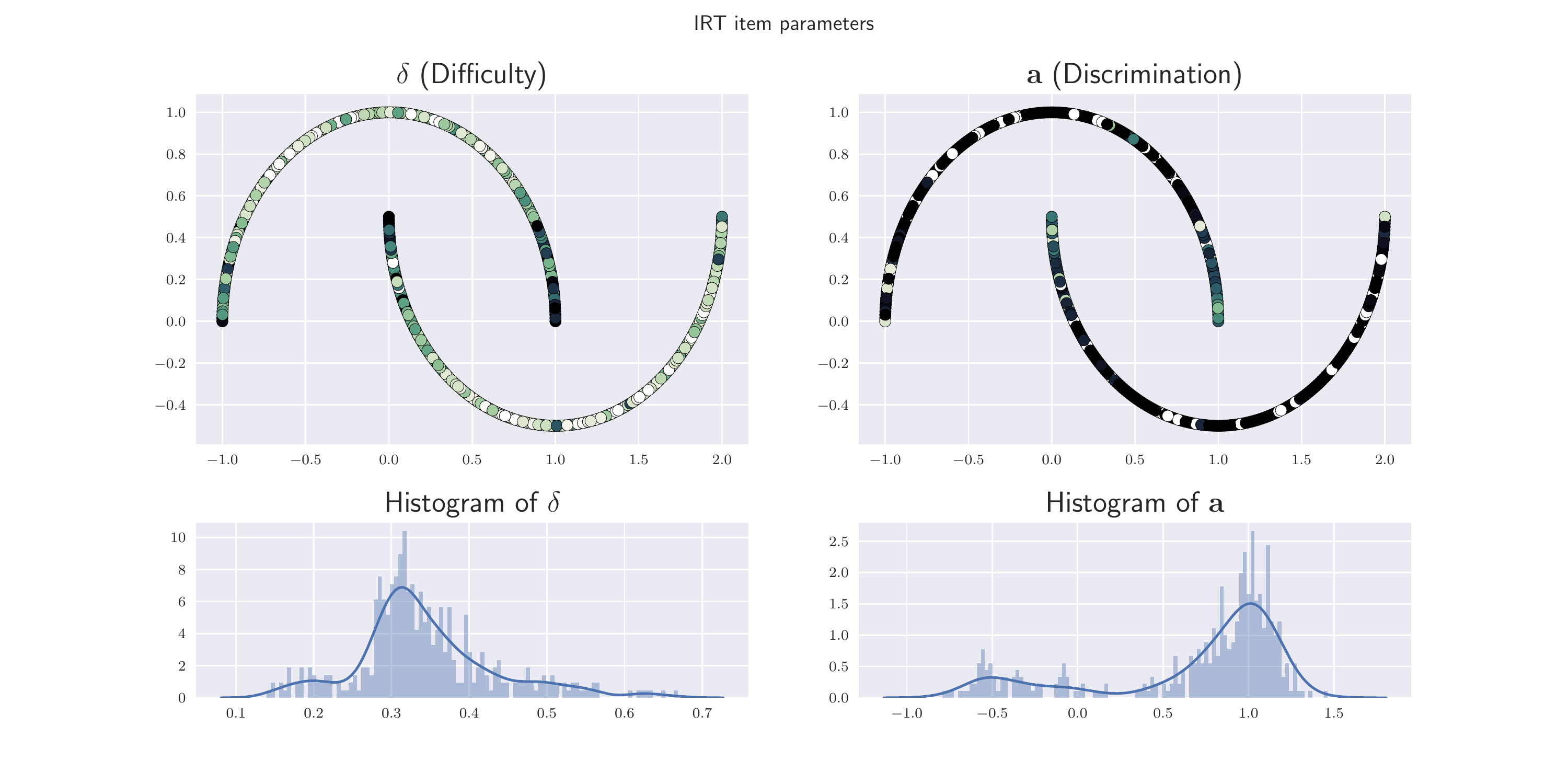}
    \caption{Dataset \moons{}.}
    \label{fig:moon_r1}
    \end{subfigure}
\caption{Inferred Latent Variables of Items of Two Synthetic Datasets. Darker colour indicates higher value. Items closer to the class boundary get higher difficulty and lower discrimination.}
\label{fig:item_params}
\end{figure*}

We now apply \BIRT{} to supervised machine learning.  
Here, each `respondent' is a different classifier and items are instances from a classification dataset\footnote{Note that an entire dataset could be considered as an item too, with associated difficulty.}. The responses are the probabilities of the \emph{correct} class assigned by the classifiers to each instance.
%
Specifically, the observed response is given by $p_{ij} = \sum_{k} \Ind(y_j=k)p_{ijk}$,
%
 where $\Ind(\cdot)$ is the indicator function, $y_j$ is the label of item $j$, $k$ is the index of each class, and $p_{ijk}$ is the predicted probability of item $j$ from class $k$ given by classifier $i$. 
The application of \BIRT{} in classification tasks aims to answer the following:
\begin{enumerate}[noitemsep]
    \item Can item parameters be used to characterise instances in terms of: (i) how difficult it is to estimate class probabilities for each instance; and (ii) how useful each instance is to discriminate between good and bad probability estimators?
    \item Does respondent ability serve as a new measure of performance that can be used to complement classifier evaluation?
\end{enumerate}
%
The following steps are adopted to obtain the responses from $M$ classifiers in a dataset:
\begin{enumerate*}
    \item Train all $M$ classifiers on a training set;
    \item Use all trained classifiers to predict the probability of each class for each data instance of a test set, which gives $p_{ijk}$;
    \item Compute $p_{ij}$ from $p_{ijk}$. 
\end{enumerate*} 
Inference is then performed in the \BIRT{} model using the responses of the $M$ classifiers to $N$ test instances. 
%
\subsection{Experimental setup}
We first applied the \BIRT{} model on two synthetic binary classification
datasets, \moons{} and \clusters{}, chosen because they are convenient for
visualisation. Both datasets are available in scikit-learn \cite{scikit-learn}.
Each dataset is divided into training and test sets, each with 400 instances. We
also tested the model on classes 3 vs 5 from the \MNIST{} dataset
\cite{lecun2010mnist}, chosen as they are similar and contain difficult
instances. For the \MNIST{} dataset, the training and test sets have 1000
instances. The classes are balanced in each dataset. We inject noise in the test
set by flipping the label $y_j$ for 20\% of randomly chosen data instances. The
hyperparameter of discrimination $\sigma_0$ is set to $1$ across all tests
unless specified explicitly. Source code can be found in
\url{https://github.com/yc14600/beta3_IRT}.

We tested 12 classifiers in this experiment:
\begin{enumerate*}[label=(\roman*)]
    \item Naive Bayes;
    \item  \ac{MLP} (two hidden layers with 256 and 64 units);
    \item AdaBoost;
    \item \ac{LR};
    \item \ac{KNN} (K=3); 
    \item \ac{LDA};
    \item \ac{QDA};
    \item Decision Tree;
    \item Random Forest;
    \item Calibrated Constant Classifier (assign probability $p=0.5$ to all instances); \label{item:const}
    \item Positive classifier (always assign positive class to instances); \label{item:pos}
    \item Negative classifier (always assign negative class to instances). \label{item:neg}
\end{enumerate*}
All except the last three are taken from scikit-learn \cite{scikit-learn} using default configuration unless specified explicitly.   

\begin{figure}[!tb]
    \centering
    \begin{subfigure}{0.75 \linewidth}
        \includegraphics[width=\linewidth,trim={1.cm 0.1cm 1.cm .5cm},clip]{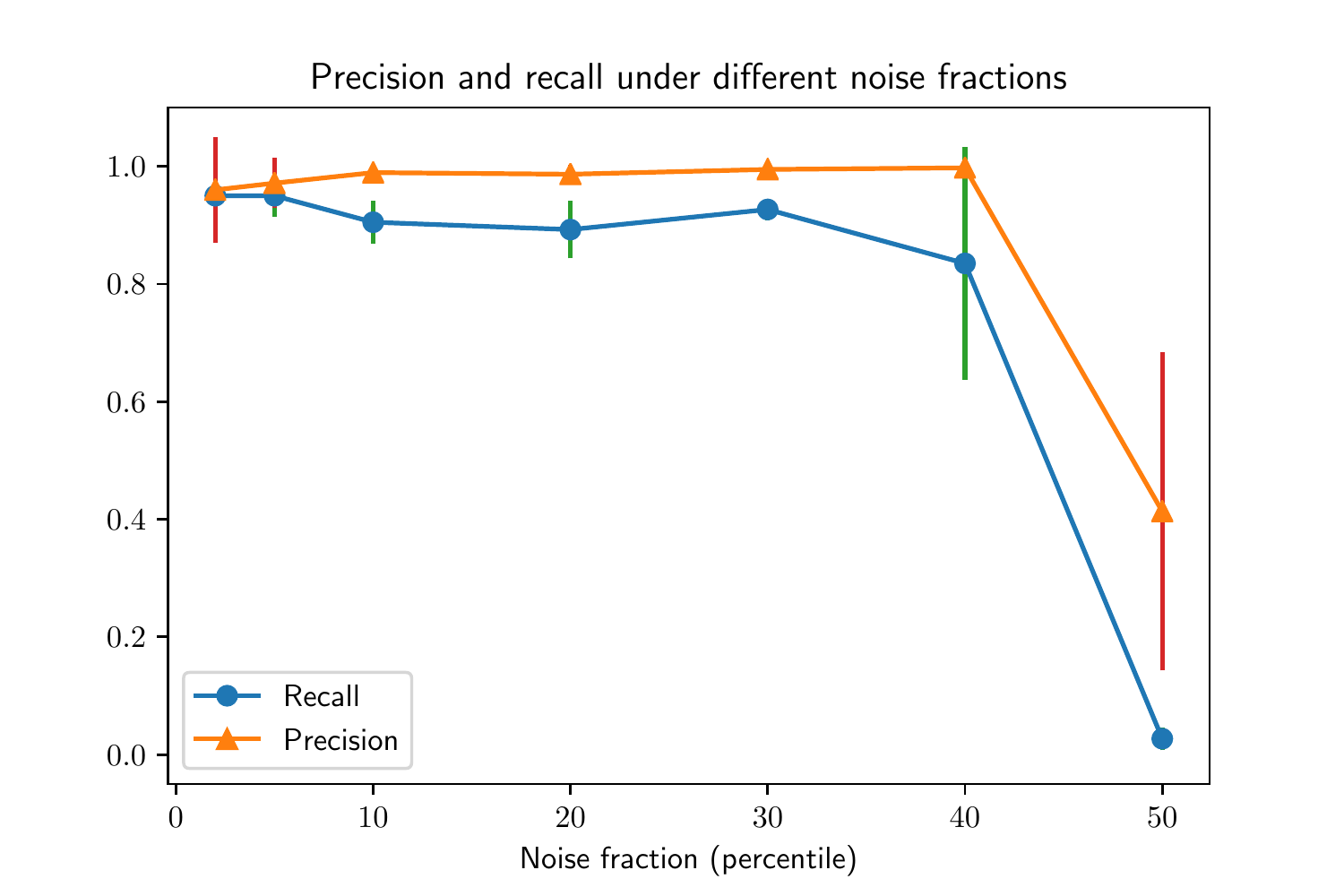}
    \caption{Noise only in test set}
    \label{fig:noise_test}
    \end{subfigure}
    \\
    \begin{subfigure}{0.75\linewidth}
        \includegraphics[width=\linewidth,trim={1.cm 0.1cm 1.cm .5cm},clip]{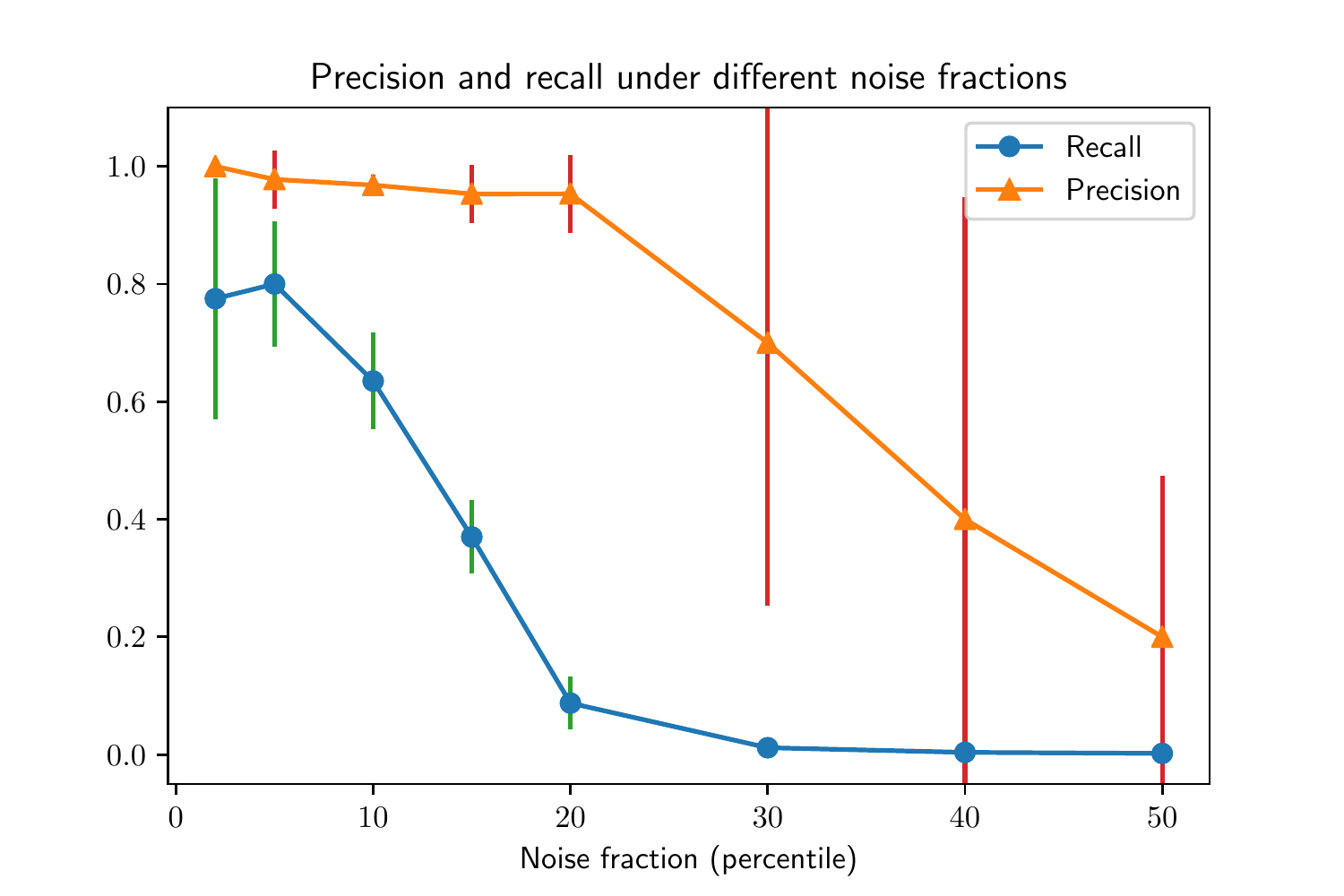}
    \caption{Noise in training and test set}
    \label{fig:noise_train}
    \end{subfigure}
\caption{Denoising Performance of Negative Discrimination in Different Settings. Tested on MNIST dataset, means and standard deviations over 5 runs of all combinations of any two classes.}
\label{fig:denoise}
\end{figure}

\subsection{Exploring item parameters}
%

\begin{figure*}[!tb]
\centering
\begin{subfigure}{0.24\linewidth}
\includegraphics[width=\linewidth,trim={0.3cm 0.2cm 0cm 0.1cm},clip]{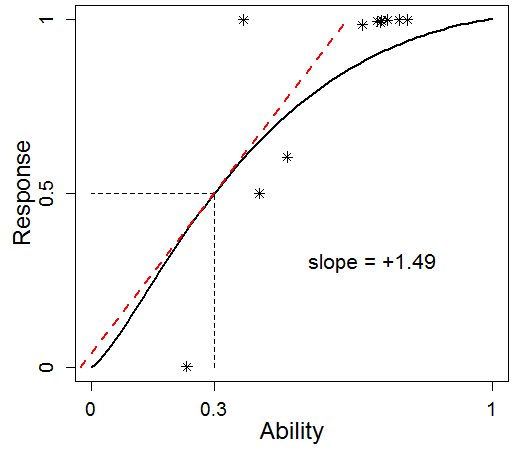}
\caption{Difficulty \low, positive discrimination \high. Instance is non-noisy item far from decision boundary.}
\label{fig:caseA_cluster}
\end{subfigure}
\hfill%
\begin{subfigure}{0.24\linewidth}
\includegraphics[width=\linewidth,trim={0.3cm 0.2cm 0cm 0.1cm},clip]{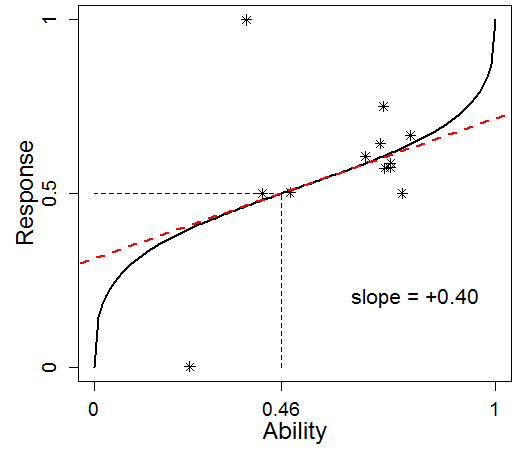}
\caption{Difficulty \high, positive discrimination \low. Instance is non-noisy item close to decision boundary.}
\label{fig:caseB_cluster}
\end{subfigure}
\hfill
\begin{subfigure}{0.24\linewidth}
\includegraphics[width=\linewidth,trim={0.3cm 0.2cm 0cm 0.1cm},clip]{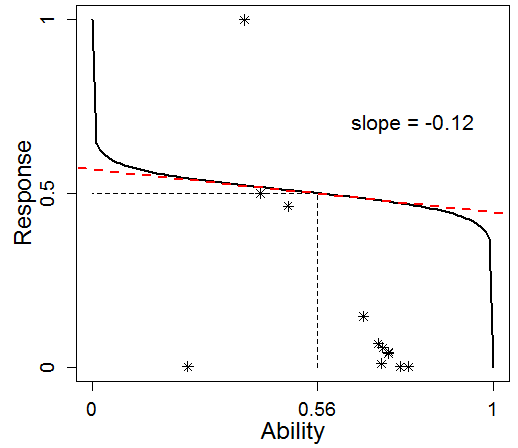}
\caption{Difficulty \high, negative discrimination \low. Instance is noisy item close to decision boundary.}
\label{fig:caseC_cluster}
\end{subfigure}
\hfill%
\begin{subfigure}{0.24\linewidth}
\includegraphics[width=\linewidth,trim={0.3cm 0.2cm 0cm 0.1cm},clip]{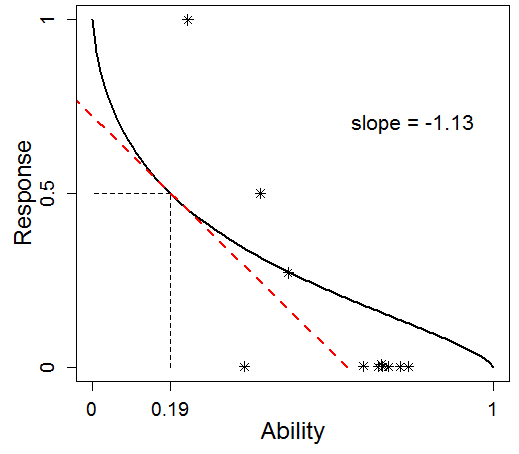}
\caption{Difficulty \low, negative discrimination \high. Instance is noisy item far from decision boundary.}
\label{fig:caseD_cluster}
\end{subfigure}
\\
\caption{Examples of ICC in the \clusters{} Dataset. Stars are the actual classifier responses fit by the ICCs.}
\label{fig:ICC_examples}
\end{figure*}

\Cref{fig:moon_r1,fig:cluster_r1} show the posterior distributions of difficulty and discrimination in the \clusters{} and \moons{} datasets. 
Instances near the true decision boundary have higher difficulties and lower discriminations for both datasets, whereas instances far away from the decision boundary have lower difficulties and higher discriminations. \Cref{fig:ICC_examples} illustrates \acp{ICC} for different combinations of difficulty, discrimination and ability. Our model provides more flexibility than logistic-shape ICC.

\begin{figure*}[!tb]
    \centering
    \begin{subfigure}{0.245\linewidth}
    \includegraphics[width=\linewidth,trim={.8cm .3cm 1.75cm 1.2cm},clip]{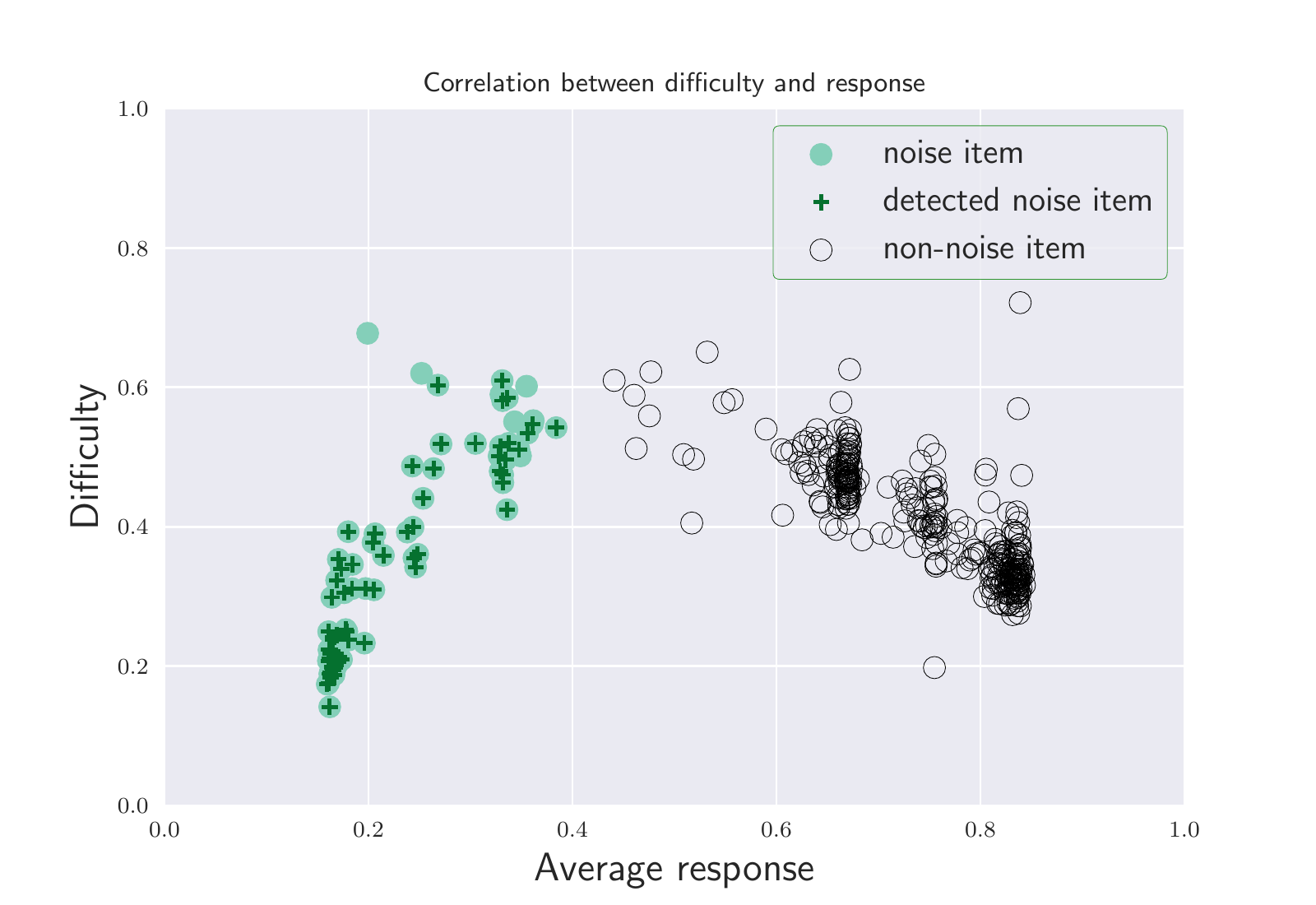}
    \caption{$\a_j$ is learned.}
    \label{fig:mnist_nf20}
    \end{subfigure}
    \begin{subfigure}{0.245\linewidth}
    \includegraphics[width=\linewidth,trim={.8cm .3cm 1.75cm 1.2cm},clip]{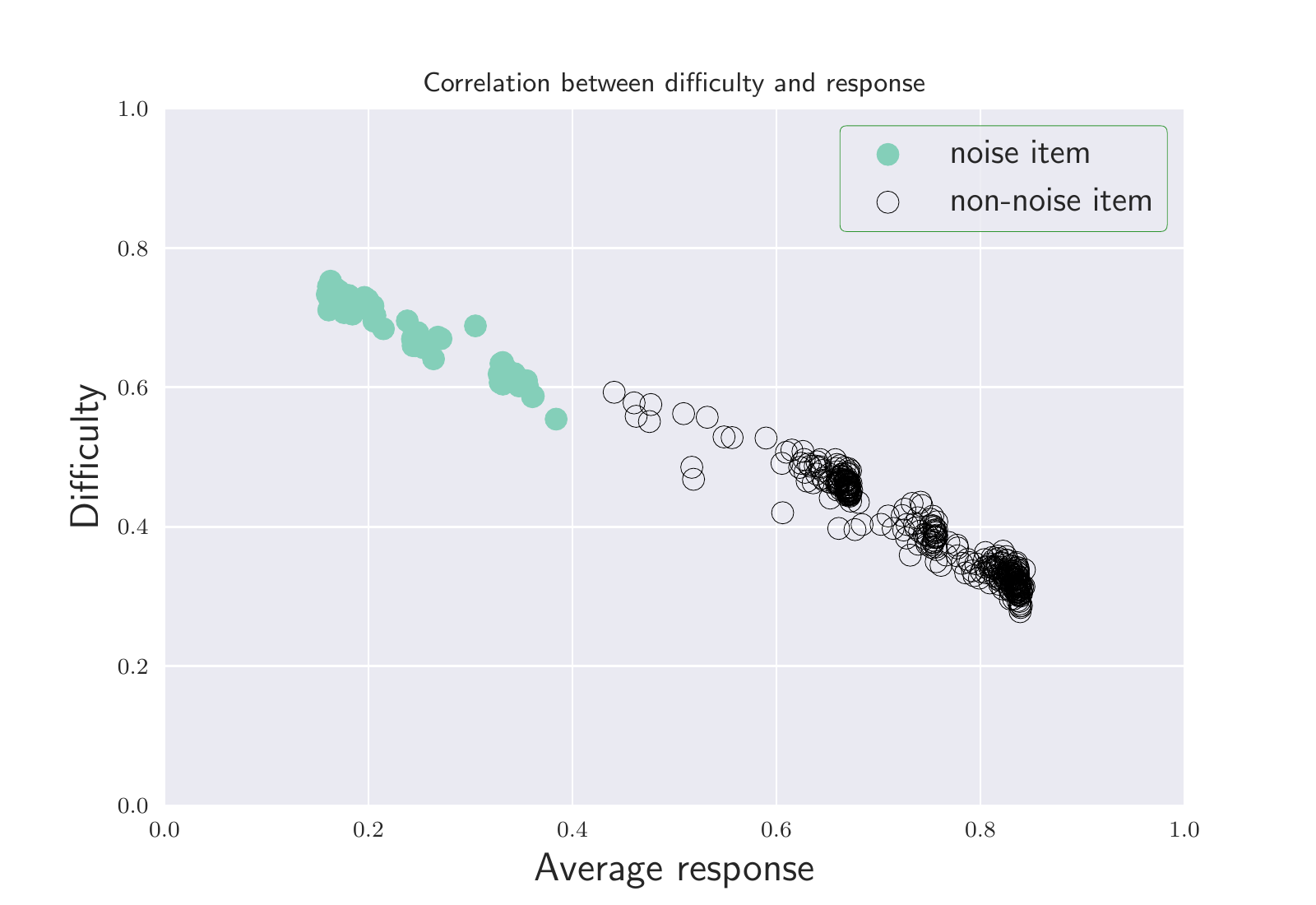}
    \caption{$\a_j$ is fixed to 1.}
    \label{fig:mnist_nf20_fixeda}
    \end{subfigure}
    \begin{subfigure}{0.245\linewidth}
    \includegraphics[width=\linewidth,trim={.8cm .3cm 1.75cm 1.2cm},clip]{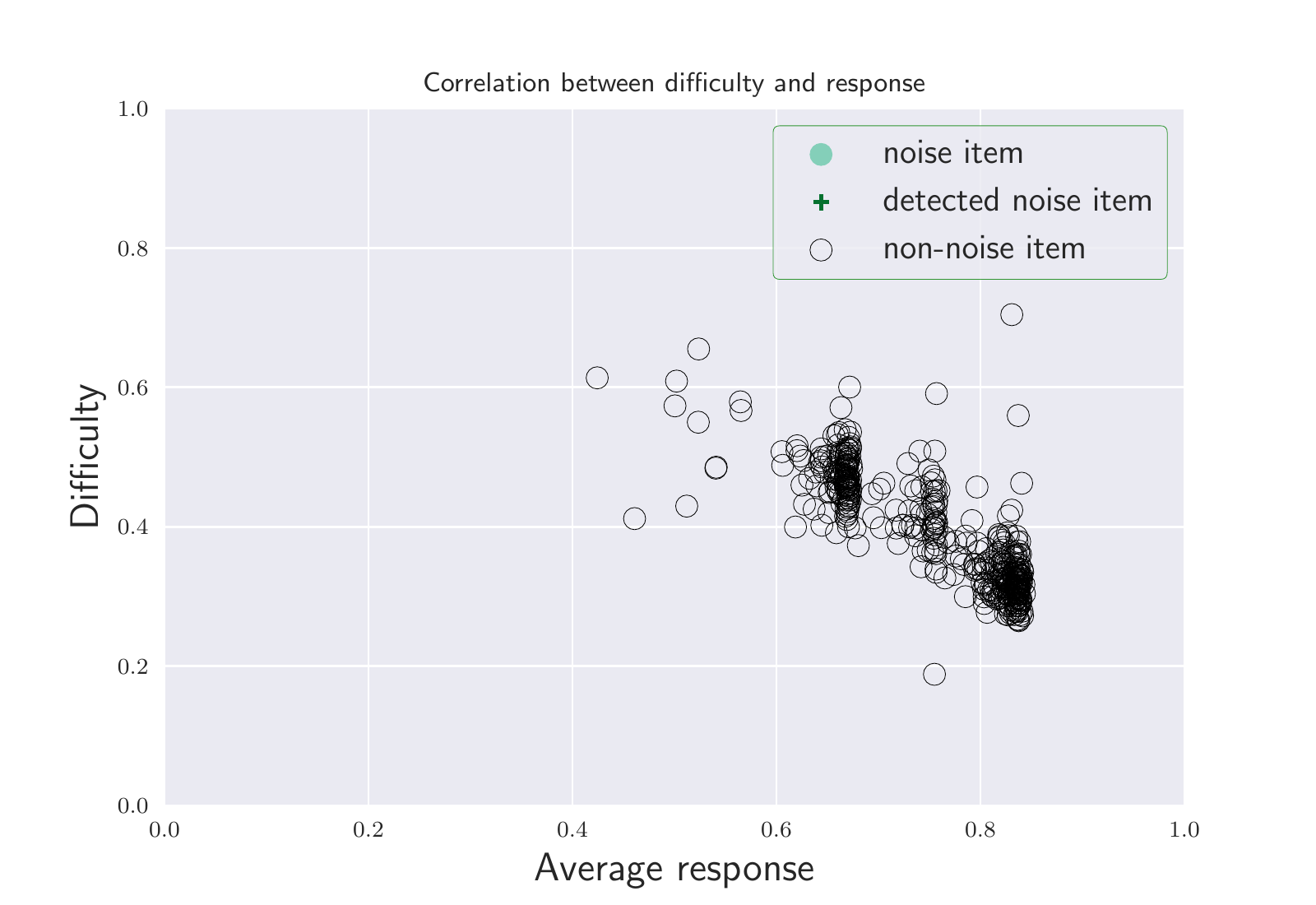}
    \caption{$\a_j$ is learned.}
    \label{fig:mnist_nf0}
    \end{subfigure}
    \begin{subfigure}{0.245\linewidth}
    \includegraphics[width=\linewidth,trim={.8cm .3cm 1.75cm 1.2cm},clip]{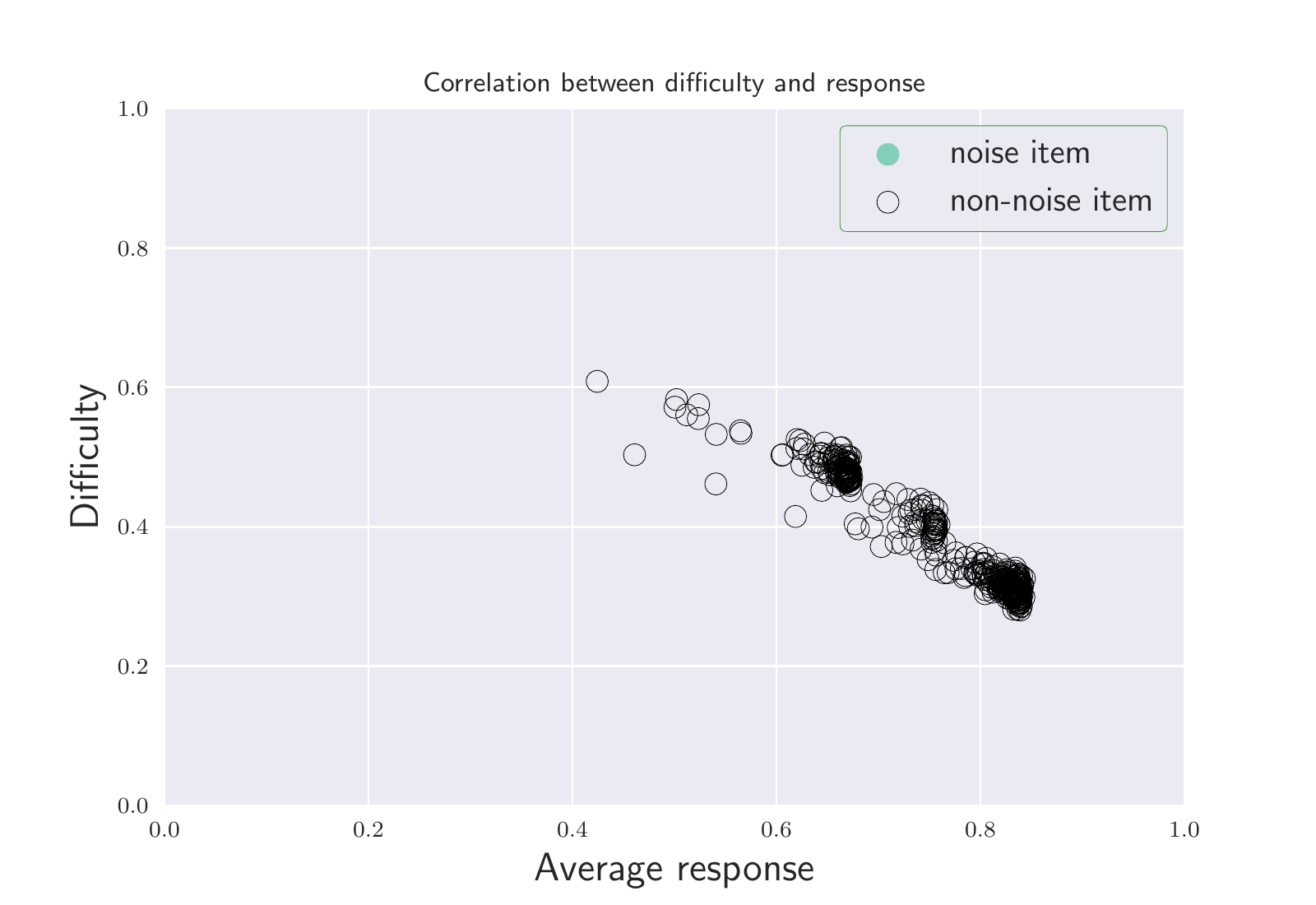}
    \caption{$\a_j$ is fixed to 1.}
    \label{fig:mnist_nf0_fixeda}
    \end{subfigure}
\caption{Correlation between Average Response and Difficulty Changes. (Under different settings of discrimination, shown for classes 3 vs 5 of \MNIST{} dataset. (a), (b) are from test data with 20\% injected noise; (c), (d) no noise.)}
\label{fig:param_corr}
\end{figure*}

There are some items inferred to have negative discrimination: these are mostly items with incorrect labels, as shown in \Cref{fig:mnist_nf20}. The negative discrimination fits the case when a low-valued response (correctness) is given by a classifier with high ability to an item with low difficulty.
\Cref{fig:mnist_nf20,fig:mnist_nf20_fixeda} shows that negative discrimination flips high difficulty to low difficulty in comparison with the results where the discrimination is fixed. \Cref{fig:mnist_nf0,fig:mnist_nf0_fixeda} show that when there are no noisy items, no negative discriminations are inferred by the model.

However, unlike the common setting of label noise detection \cite{frenay2014classification}, using negative discrimination to identify noisy labels requires that the training set can only include very few noisy examples. The reason is that noise in the training set introduces noise to all classifiers' abilities, and hence the noise in test set is hard to be identified. The experiment results of such cases are compared in \Cref{fig:denoise}. This is a common issue in ensemble-based approaches for noise detection, which has been addressed for instance in \cite{sluban2015relating}. Our model can be trained on a small noise-free training set and then updated incrementally with identified non-noisy items, which is still practical in real applications.
In contrast, in students answer experiments, there is no separate training set to build students' abilities before getting their answers for questions because the students are assumed to be trained by formal education already.

\subsection{Assessing the ability of classifiers}



\begin{figure}[!tb]
     \centering
     \includegraphics[width=0.7\linewidth]{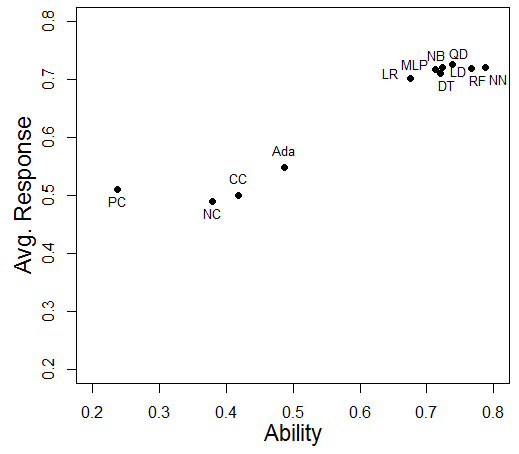}
     \caption{Ability vs Average Response in the \clusters{} dataset. The classifiers in the top right getting similar avg, response around 0.7, but their abilities are diverse from 0.65 to 0.8.}
     \label{fig:ability_response}
 \end{figure}

\Cref{fig:ability_response} shows a linear-like relation between ability and average response except the top right and bottom left of the figure. However, most classifiers are in the non-linear part, with ability between 0.7 and 0.8 and avg. response around 0.72, and the highest ability does not correspond to the highest avg response. This is caused by the element-wise difference which we will discuss below.

\Cref{tab:measures} shows the comparison between abilities and several popular classifier evaluation metrics on the \MNIST{} dataset, while \Cref{tab:spearman} gives the Spearman's rank correlation between these metrics. The experiment results of \clusters{} are provided in the supplementary materials. We can see that ability behaves differently from the other metrics since it is not only estimated using aggregates of predicted probabilities, but also by the difficulties and discriminations of corresponding items. This can be seen from the equation below:
\begin{equation}\label{eq:ability_analysis}
   \left(\frac{1}{\bar{p}_{ij}} - 1\right)^{\a_j^{-1}} \left(\frac{1}{\deltab_j} - 1\right) = \frac{1}{\thetab_i} - 1,\;
   \bar{p}_{ij} = \frac{\alpha_{ij}}{\alpha_{ij}+\beta_{ij}}
\end{equation}
Where $\bar{p}_{ij}$ is the expected response of item $j$ given by classifier $i$, $\alpha_{ij}$ and $\beta_{ij}$ are defined in \Cref{eq:model_def}. For example, a low $\bar{p}_{ij}$ for a difficult instance will not give high penalty to the ability $\thetab_i$ because the difficulty $\deltab_j$ is high and the discrimination $\a_j$ is often close to zero for difficult items as we observed in \Cref{fig:item_params}, whereas log-loss will generate high penalty as long as the correctness is low and Brier score will only consider the correctness as well. The ability learned by our model provides a new scaled measurement for classifiers, which evaluates the performance of probability estimation in a sense of weighted instance-wise basis.

\begin{figure}[!tb]
     \centering
     \includegraphics[width=.8\linewidth,trim={.5cm .2cm .5cm .5cm},clip]{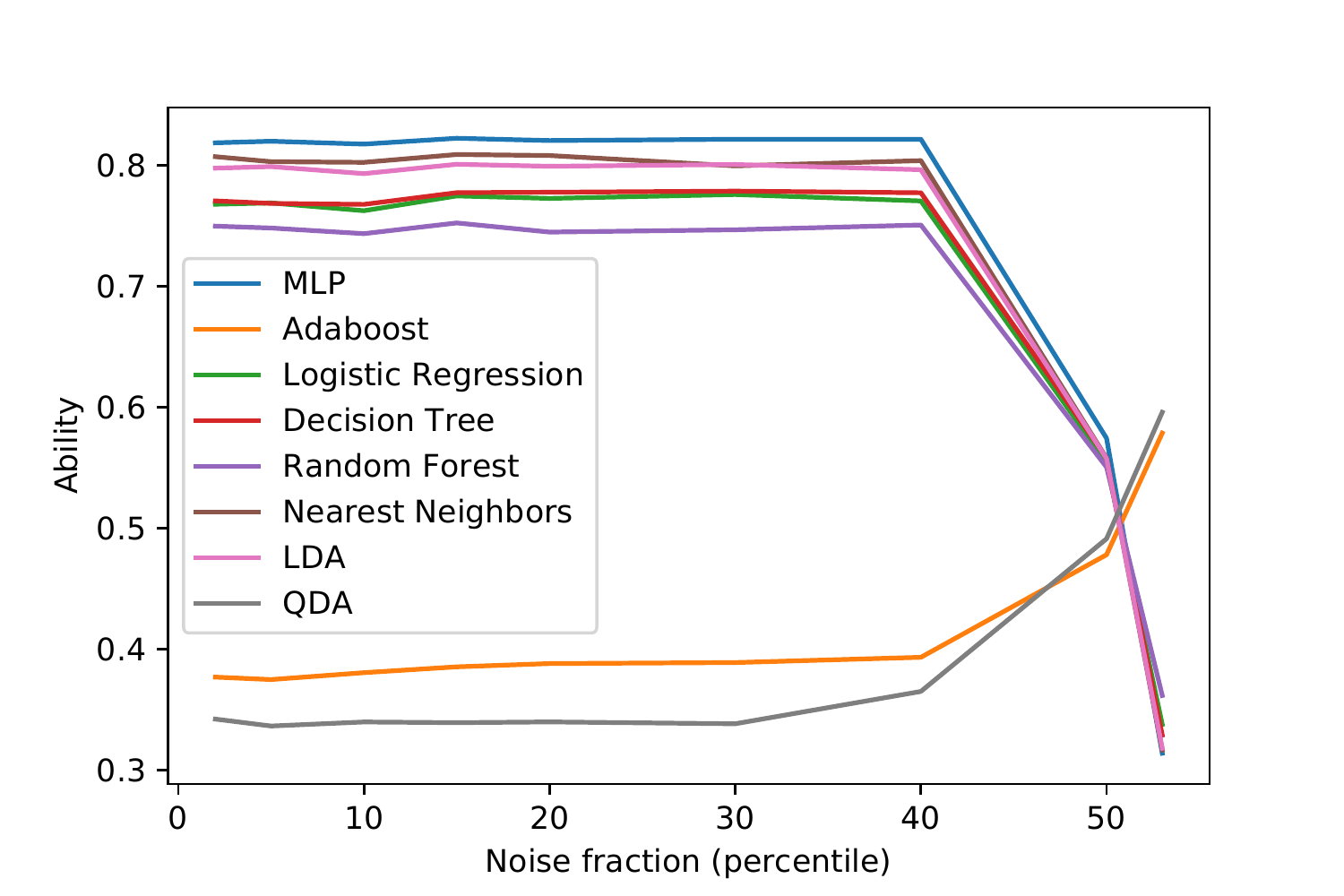}
     \caption{Ability of classifiers with different noise fractions of test data, which shows the ability is robust to noisy test data.}
     \label{fig:ability_noise}
 \end{figure}

Another advantage of ability as a measurement of classifiers is that it is robust to noisy test data. \Cref{fig:ability_noise} demonstrates that the inferred abilities of the classifiers stay nearly constant as the noise fraction in the test set is increased until half of the test points are incorrectly labelled.

\begin{table}[!tb]
\setlength{\tabcolsep}{2pt}
\caption{Comparison between Ability and other Classifier Performance Metrics (MNIST)}
\vspace{5.5pt}
\centering
\scriptsize
\label{tab:measures}

\begin{tabular}{cccccccc}
\toprule
                    & Avg. Resp. & Ability & Accuracy & F1 score & Brier score &Llog loss & AUC \\
\midrule
DT       & 0.7398            & 0.7438  & 0.7425   & 0.7297    & 0.2337      & 1.1537   & 0.7776          \\
NB         & 0.6439            & 0.7423  & 0.6425   & 0.6951    & 0.3533      & 10.6097  & 0.6799          \\
MLP                 & 0.7826            & 0.8384  & 0.7825   & 0.774     & 0.2086      & 2.457    & 0.7951          \\
Ada.            & 0.5621            & 0.4887  & 0.775    & 0.7656    & 0.2036      & 0.5959   & 0.79            \\
RF       & 0.7215            & 0.7395  & 0.7725   & 0.7573    & 0.1926      & 3.4979   & 0.8119          \\
LDA                 & 0.7185            & 0.8052  & 0.72     & 0.7098    & 0.2714      & 5.1328   & 0.7276          \\
QDA                 & 0.5948            & 0.5892  & 0.595    & 0.6611    & 0.405       & 13.9884  & 0.5854          \\
LR & 0.7699            & 0.8001  & 0.7775   & 0.7688    & 0.2059      & 1.4246   & 0.7939          \\
KNN   & 0.7645            & 0.8228  & 0.7675   & 0.7572    & 0.2111      & 6.3816   & 0.8011  \\
\bottomrule
\end{tabular}
\end{table}

\begin{table}[!tb]
\setlength{\tabcolsep}{2pt}
\caption{Spearman's Rank Correlation between Ability and other Classifier Performance Metrics (MNIST)}
\vspace{5.5pt}
\centering
\scriptsize
\label{tab:spearman}
\begin{tabular}{cccccccc}
\toprule
                  & Avg. Resp. & Ability & Accuracy & F1 & Brier & Log loss & AUC \\
\midrule
Avg. Resp. & 1.0               & 0.8333  & 0.6      & 0.6       & 0.2833       & 0.2       & 0.6             \\
Ability           & 0.8333            & 1.0     & 0.3333   & 0.3333    & -0.05        & -0.05     & 0.35            \\
Accuracy          & 0.6               & 0.3333  & 1.0      & 1.0       & 0.8333       & 0.7       & 0.75            \\
F1         & 0.6               & 0.3333  & 1.0      & 1.0       & 0.8333       & 0.7       & 0.75            \\
Brier      & 0.2833            & -0.05   & 0.8333   & 0.8333    & 1.0          & 0.6833    & 0.8333          \\
Log loss         & 0.2               & -0.05   & 0.7      & 0.7       & 0.6833       & 1.0       & 0.3667          \\
AUC   & 0.6               & 0.35    & 0.75     & 0.75      & 0.8333       & 0.3667    & 1.0    \\
\bottomrule
\end{tabular}
\end{table}

\section{CONCLUSIONS}
\label{sec:conclusion}

This paper proposed \BIRT{}, a new IRT model solves the limitations of a previous continuous IRT approach \cite{noel2007beta}, by adopting a new formulation which allows the model to produce a new family of \aclp{ICC} including sigmoidal and anti-sigmoidal curves. Therefore, our \BIRT{} model is more versatile than previous IRT models, being able to model a significantly more expressive class of response patterns. Additionally, our new formulation assumes bounded support for abilities and difficulties in the $[0,1]$ range which produces more natural and interpretable results than previous IRT models.

We evaluated \BIRT{} in two experimental scenarios. First, \BIRT{} was applied to the psychometric task of student performance estimation, 
\BIRT{} outperformed 2PL-ND in all 33 datasets, showing the importance of the versatility of the \acp{ICC} that are produced by our approach.
We then applied \BIRT{} in a binary classification scenario.
The results showed that item parameters inferred by the \BIRT{} model can provide useful insights for difficult or noisy instances, and the inferred latent ability variable serves to evaluate classifiers on an instance-wise basis in terms of probability estimation. To the best of our knowledge, this was the first time that an IRT model was used for these tasks.

Future work includes using the model as a tool for model selection or ensembling. 
Another use is to design datasets for benchmarking, based on estimated difficulties and discriminations. 
Another extension would be to replace the Beta with a Dirichlet distribution to cope with multi-class scenarios.

\subsubsection*{Acknowledgements}
Part of this work was supported by The Alan Turing Institute under EPSRC grant EP/N510129/1.
Ricardo Prudêncio was financially supported by CNPq (Brazilian Agency).
\bibliography{references}
\end{document}